\documentclass[11pt]{article}
\usepackage[T1]{fontenc}
\usepackage[utf8]{inputenc}
\usepackage{lmodern}
\usepackage[a4paper,margin=25mm]{geometry}
\usepackage{amsmath,amssymb}
\usepackage{booktabs}
\usepackage{graphicx}
\usepackage{xcolor}
\usepackage{microtype}
\usepackage[numbers,sort&compress]{natbib}
\usepackage{tikz}
\usetikzlibrary{arrows.meta,positioning,fit}
\usepackage[hidelinks]{hyperref}
\usepackage{xurl}

\newcommand{\tx}{\ensuremath{\times}}

\title{Working Around the Compute Ceiling:\\
Galahad's Byte-Exact Memory Makes\\
LLM Reading a One-Time Cost}
\author{Sietse Schelpe\\
Corbenic AI\\
\texttt{sietse@corbenic.ai}}
\date{October 2026\\[6pt]
{\footnotesize \copyright\ 2026 Corbenic AI. Licensed under CC BY 4.0
(\url{https://creativecommons.org/licenses/by/4.0/}).\\
Any reuse must credit: Sietse Schelpe, Corbenic AI (2026).}}

\begin{document}
\maketitle

\begin{abstract}
A transformer language model performs a bounded amount of computation per
token, and recent work by Vishal Sikka, former CEO of Infosys, argues that this
bound limits which tasks a model can carry out or verify
\citep{sikka-hallucination-2025}. We do not try to raise
that ceiling. We ask how much of the budget beneath it is spent on work the
model has already done. Serving is stateless across requests: a model that
answers a second question about a document recomputes the document's attention
state from the first token. On seven real-world datasets, 98.7\% of prompt
tokens were text the model had already read. We present Galahad, a memory layer
for vLLM, SGLang and llama.cpp that makes this reading a one-time cost.
Taliesin saves the model's key--value (KV) state for a block of text and loads
it on the next request that contains the same bytes, instead of recomputing
it. Blaise keeps the documents themselves and passes the model only the section
a question needs. We measure each part separately on a recall test with 100
facts hidden in a 97{,}000-token corpus (Gemma~4 31B). Taliesin alone let the
model attend to the whole corpus and answered 98 of 100 on llama.cpp at 3.0\,s
and 572\,J per question, against 10 of 100, 9.3\,s and 2{,}754\,J for the same
model without Galahad, which could hold only the last 12{,}000 tokens. With
Blaise added, the model read about 668 tokens per question and answered 100 of
100 on all three runtimes at 0.59--0.64\,s and 200--213\,J; a tuned RAGFlow
pipeline answered 77. Storing the corpus is a one-time cost of about 100\,s
and 28\,kJ, whose energy is recovered after 13 questions.
Restored state is bit-identical to freshly computed state: all 262{,}144
output logits matched after restart, rehydration and hot-load. Galahad worked with
all 30 models we tested (30/30) under vLLM. Galahad fails closed: any load
that does not pass its checks is recomputed, so memory never changes an answer.
Together these results move LLM serving from stateless to stateful inference.
Galahad is available as a free,
non-commercial beta for one GPU at \url{https://github.com/corbenicai/galahad}.
\end{abstract}

\section{Introduction}
\label{sec:intro}

Varin Sikka and Vishal Sikka, the former CEO of Infosys
\citep{sikka-hallucination-2025}, argue that a transformer
performs $O(N^2 \cdot d)$ computation for a prompt of $N$ tokens and model
dimension $d$, and that a task whose complexity exceeds this bound cannot be
carried out correctly by the model, or verified by it. This paper accepts that
ceiling. It addresses a different question: how much of the computation below
the ceiling is spent on useful work, and how much on repeating work the model
has already done.

A large language model reads before it writes. For every request, the serving
engine runs the prompt through the model once (prefill) to build the attention
keys and values that the generated tokens attend to. For long prompts, prefill
dominates the time to first token and a large share of the energy spent per
request.

In most applications the same text is read many times. A support assistant
answers hundreds of questions against the same manuals. A coding agent re-sends
the same repository files on every step. A contract review tool asks dozens of
questions of one agreement. Current engines reuse attention state only within
a narrow scope: vLLM and SGLang keep a prefix cache in GPU memory
\citep{kwon-pagedattention-2023,zheng-sglang-2024}, which is evicted under
memory pressure and lost when the process restarts. Offloading systems extend
this to CPU memory and disk \citep{liu-lmcache-2025,qin-mooncake-2025}. In the
common case, however, a document that the model read yesterday is read again
from the first token today.

We argue that this should change: \emph{the unit of inference cost should be
new text, not total text}. A model should pay to read a document once, and
every later question about it should cost only the question and the answer.
This moves inference from a stateless computation to one with persistent
memory, in the same way that databases separated storing data from computing
over it. The rest of this paper describes a system built on that principle and
reports what it achieves.

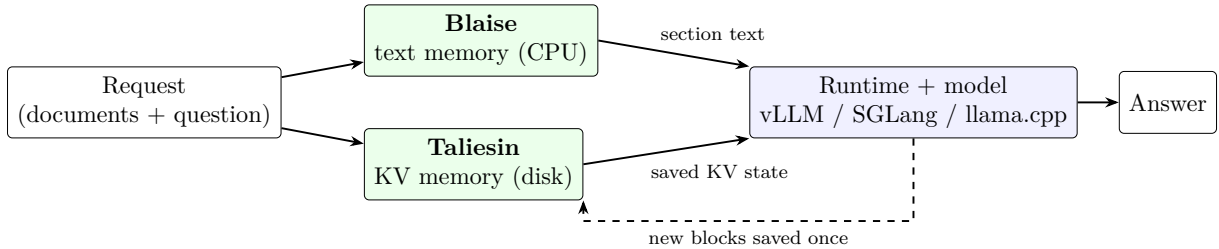
\begin{figure}[t]
\centering
\resizebox{\linewidth}{!}{\begin{tikzpicture}[
  box/.style={draw, rounded corners=2pt, align=center, font=\small, minimum height=9mm, inner sep=4pt},
  mem/.style={box, fill=green!8},
  gpu/.style={box, fill=blue!6},
  arr/.style={-{Stealth[length=2mm]}, thick},
  lab/.style={font=\scriptsize, align=center}
]
\node[box] (q) {Request\\(documents + question)};
\node[mem, right=12mm of q, yshift=9mm] (b) {\textbf{Blaise}\\text memory (CPU)};
\node[mem, right=12mm of q, yshift=-9mm] (t) {\textbf{Taliesin}\\KV memory (disk)};
\node[gpu, right=22mm of b, yshift=-9mm, minimum width=30mm] (m) {Runtime + model\\vLLM / SGLang / llama.cpp};
\node[box, right=6mm of m] (a) {Answer};
\draw[arr] (q) -- (b);
\draw[arr] (q) -- (t);
\draw[arr] (b.east) -- node[lab, above right, pos=0.35] {section text} (m.north west);
\draw[arr] (t.east) -- node[lab, below right, pos=0.35] {saved KV state} (m.south west);
\draw[arr] (m) -- (a);
\draw[arr, dashed] (m.south) |- node[lab, pos=0.75, below] {new blocks saved once} ([yshift=-3mm]t.south east) -- (t.south east);
\end{tikzpicture}}
\caption{Galahad in the serving path. Blaise selects the section of text a
question needs; Taliesin supplies saved KV state for any block the model has
already processed, and saves new blocks after their first prefill. If a load
fails any check, the runtime recomputes the block.}
\label{fig:arch}
\end{figure}

Galahad has two memories (Figure~\ref{fig:arch}):

\begin{itemize}
  \item \textbf{Taliesin} stores the model's own KV state for a block of text
  and loads it when the same bytes appear again. Loading is exact: the restored
  state produces bit-identical logits to a fresh prefill.
  \item \textbf{Blaise} stores the text itself, organised so that a question
  can be answered from one section. It runs on the CPU and passes the model the
  exact text of that section, not the whole corpus.
\end{itemize}

Taliesin removes repeated computation; Blaise removes unnecessary reading.
Each can run without the other.

This paper makes four contributions:

\begin{enumerate}
  \item A memory layer that plugs into three production runtimes (vLLM,
  SGLang, llama.cpp) as one shared library, and that falls back to normal
  computation on any load failure (Section~\ref{sec:system}).
  \item Evidence that restored state is exact: bit-identical logits after
  save and restore, and byte-equal prefill across four model architectures
  (Section~\ref{sec:correctness}).
  \item Measurements of accuracy, latency and GPU energy for each part
  separately, on a controlled recall test and on seven real-world datasets,
  including an external retrieval baseline, and coverage of 30 of 30 tested
  models (Section~\ref{sec:eval}).
  \item The design boundaries of the approach and why they follow from how
  transformers and serving engines work (Section~\ref{sec:limits}).
\end{enumerate}

This is a system description. It does not describe the storage format, the
internal structure of Blaise, or the security design in implementation detail;
these are covered by pending patent applications.

\section{System}
\label{sec:system}

\subsection{Deployment}
Galahad ships as one shared library (\texttt{libgalahad.so}) for Linux x86-64
with two runtime dependencies (OpenSSL's libcrypto and zstd). It connects to
vLLM as a KV connector, to SGLang as a HiCache storage backend, and to
llama.cpp through slot save and restore. No model weights or runtime source
code are modified. The library refuses to start without a tenant identity and
a model fingerprint, so state saved by one model or tenant cannot be loaded by
another.

\subsection{Taliesin: KV memory}
After the runtime computes the KV state for a block of prompt text, Taliesin
writes it to storage under a fingerprint of the exact input bytes, the model
and the tenant. When a later request contains the same block, the runtime
loads the stored state instead of recomputing it. Stored state is checked on load and can be encrypted at rest. The design rule is \emph{serve without memory, never
with wrong memory}: if a record is missing, damaged, or fails a check, the
runtime recomputes the block and the request completes normally.

Reuse happens at block level, where token positions match exactly
(Section~\ref{sec:limits}).

\subsection{Blaise: text memory}
Blaise keeps each document as byte-exact text, divided into sections. For a
question, it selects one section on the CPU and passes that section's text to
the model, which then reads it and answers. In the recall test of
Section~\ref{sec:recall}, Blaise passed about 668 tokens per question instead
of the full 97{,}000-token corpus. We do not describe its selection method here.

A second mode, in which the model itself reads Blaise's index of the corpus and
Taliesin keeps that reading so that it is paid once, is part of the design.
This paper reports the first mode.

\section{Correctness}
\label{sec:correctness}

A memory layer is useful only if loaded state is the state the model would
have computed. Table~\ref{tab:correct} summarises the checks.

\begin{table}[t]
\centering
\small
\caption{Correctness checks. Each row is a separate experiment.}
\label{tab:correct}
\begin{tabular}{@{}p{4.4cm}p{4.1cm}p{5.6cm}@{}}
\toprule
Check & Result & Conditions \\
\midrule
Save and restore & 262{,}144 of 262{,}144 logits bit-identical; 27 checks &
After process restart, rehydration from disk and hot-load. Gemma~4 12B Q8\_0,
llama.cpp, 2026-08-22 \\
Deterministic prefill & 20 of 20 byte-equal (SHA-256 of logits) &
Two fresh prefills of the same 1{,}024-token prompt compared. Llama~3.1 70B,
Gemma~4 E4B, Qwen~3.6 35B-A3B, Ministral~3 8B; H200, llama.cpp b9189 \\
State moved across GPU types & 64 of 64 greedy tokens identical &
KV state saved on A6000, loaded on RTX 4090 and the reverse \\
Stored bytes across GPUs & 3 of 3 on each of 10 GPU types & Same record
written and read back \\
Hybrid architecture & 5 of 5 token-exact against Galahad-off reference &
Qwen3.5 9B (Gated DeltaNet), tensor parallel 2 \\
Snapshots & 276 of 276 tokens bit-identical & After restart; RTX PRO 4500 \\
Encryption round trip & 9.77\,GiB, 0 failures; NIST CAVP 4{,}167 vectors, 0
failures & T4 and A40 \\
\bottomrule
\end{tabular}
\end{table}

Exactness matters because approximate reuse fails silently. A prefix cache
that returns slightly different state can change generated tokens without any
error, as reported for AMD MI355X GPUs in the vLLM issue tracker
\citep{vllm-issue-33123-2025}. Floating-point non-determinism in batched
inference is a known source of such differences \citep{he-nondeterminism-2025}.
Galahad treats a mismatch as a failed load and recomputes.

\paragraph{Sabotage testing.}
For each safety mechanism (confirmation hash, licence signature, tenant
separation, byte comparison on load), we ran the corresponding attack three
times: with the defence on, with it switched off, and with it restored. A
defence counts as effective only if the attack succeeds when it is off and
fails when it is on. All four defences met this criterion. A separate
pass with ten analysis tools, including address sanitizers, found one
out-of-bounds read, which we fixed before release.

\section{Evaluation}
\label{sec:eval}

\paragraph{Method.}
All benchmarks ran on rented cloud GPUs; local results were not counted. No
configuration was told where an answer was located. Every retry and every
loaded block counts toward time, energy and token totals. Time is the median
per question. Energy is the total from the GPU's own counter (NVML), averaged
per question; idle power was measured separately and is included. Decoding was
greedy (temperature~0).

\subsection{Recall in a long corpus}
\label{sec:recall}

\paragraph{Setup.}
We inserted 100 facts with invented names and random numbers into 13 Wikipedia
articles (434\,KB, 96{,}726 tokens in 11 blocks) and asked for each fact in
paraphrase, at depths from 1\% to 99\% of the corpus. The model was Gemma~4 31B
(4-bit) on an RTX A6000, 26--27 September 2026. Unless marked otherwise, the
runtime's own cache was emptied before every question, so any reuse had to come
from Galahad. The baseline without Galahad used a 12{,}000-token context
window, so it saw only the last 12\% of the corpus. As an external baseline we
ran RAGFlow v0.27.2 \citep{ragflow-2026}, an open-source retrieval-augmented
generation system, with lightly tuned retrieval settings on vLLM on an L40S
GPU.

We report the three configurations separately so that the contribution of
each part is visible: Taliesin alone (the model reads the whole corpus, loaded
from memory), Taliesin with Blaise (the model reads one selected section), and
no Galahad.

\begin{table}[t]
\centering
\small
\caption{Recall of 100 hidden facts in a 96{,}726-token corpus (Gemma~4 31B,
4-bit, RTX A6000). Median seconds and mean GPU joules per question, all retries
included. The runtime's own cache was emptied before every question except in
the rows marked ``runtime cache kept''.}
\label{tab:recall}
\begin{tabular}{@{}llrrr@{}}
\toprule
Runtime & Configuration & Correct & s/question & J/question \\
\midrule
llama.cpp & No Galahad (last 12k tokens) & 10 & 9.25 & 2{,}754 \\
          & Taliesin only$^{a}$ & 98 & 3.01 & 572 \\
          & Taliesin + Blaise$^{c}$ & 100 & 0.64 & 213 \\
\midrule
vLLM      & No Galahad (last 12k tokens) & 10 & 8.11 & 2{,}402 \\
          & Taliesin only & 99 & 15.56 & 4{,}043 \\
          & Taliesin + Blaise$^{c}$ & 100 & 0.59 & 200 \\
\midrule
SGLang    & No Galahad (last 12k tokens) & 10 & 10.29 & 2{,}940 \\
          & Taliesin only$^{b}$ & 100 & 22.02 & 4{,}687 \\
          & Taliesin + Blaise$^{c}$ & 100 & 0.64 & 210 \\
\midrule
vLLM, runtime cache kept & No Galahad (last 12k tokens) & 10 & 0.14 & 93 \\
          & Taliesin only & 99 & 1.09 & 329 \\
\midrule
vLLM (L40S) & RAGFlow v0.27.2, tuned$^{d}$ & 77 & -- & -- \\
\bottomrule
\end{tabular}

\vspace{2pt}
\begin{minipage}{0.95\linewidth}\footnotesize
$^{a}$Repeated once: 98 correct, 3.00\,s, 580\,J.
$^{b}$Older SGLang backend; treat the time as an upper bound.
$^{c}$Blaise was developed on this test. On seven datasets it had never
seen, it answered 91--92\% (Section~\ref{sec:dirty}).
$^{d}$Retrieval settings lightly tuned on this test; time and energy not reported.
\end{minipage}
\end{table}

\begin{figure}[t]
\centering
\includegraphics[width=0.85\linewidth]{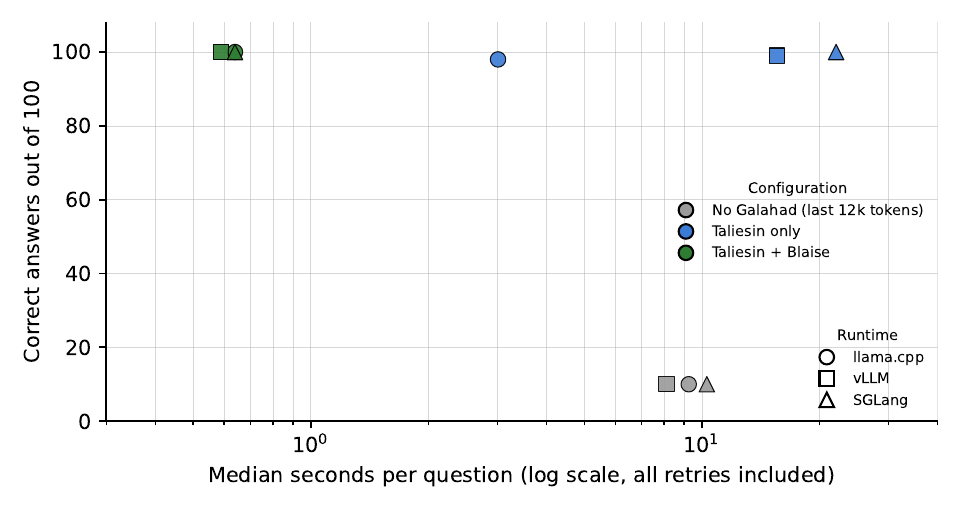}
\caption{Accuracy against median time per question for the recall test, with
the runtime's own cache emptied before every question
(Table~\ref{tab:recall}). Colour marks the configuration, shape the runtime.
Without Galahad the model cannot see 88\% of the corpus.}
\label{fig:test1}
\end{figure}

\paragraph{Taliesin: the whole corpus, loaded instead of recomputed.}
With Taliesin alone, the model attends to the full corpus and answered 98 to
100 of 100 questions on the three runtimes, against 10 without Galahad. On
llama.cpp, each question sent 53{,}219 prompt tokens on average, of which
99.5\% were loaded from Taliesin. The model therefore processed 5.5\tx{} as many
prompt tokens as the truncated baseline (9{,}700 per question) and still
finished 3.1\tx{} faster (3.01\,s against 9.25\,s) with 79\% less GPU energy
(572\,J against 2{,}754\,J). This result involves no retrieval and no tuning on
the test questions; it is the effect of KV persistence alone. With vLLM's own
cache kept between questions, Taliesin alone answered 99 of 100 at 1.09\,s and
329\,J per question.

\paragraph{Blaise: reading less.}
With Blaise, the model read one section of about 668 tokens per question and
answered all 100 questions on every runtime. On vLLM this took 0.59\,s and
200\,J per question, against 8.11\,s and 2{,}402\,J for the baseline without
Galahad: 13.7\tx{} less time and 92\% less GPU energy, while answering 100
questions instead of 10. The two memories add up: on llama.cpp, Taliesin alone
was 3.1\tx{} faster than the baseline, and Blaise, by giving the model 668
tokens instead of 96{,}726, took that to 14.5\tx{}.

\paragraph{One-time cost.}
Storing the corpus in Taliesin took one prefill of 96{,}726 tokens: 96\,s and
27.8\,kJ on llama.cpp, 108\,s and 28.4\,kJ on vLLM. On llama.cpp each later
question saved 2{,}182\,J and 6.2\,s against the baseline without Galahad, so
the stored corpus paid back its energy after 13 questions and its time after 16.
Indexing the corpus in Blaise took 0.1\,s and computed no tokens on the GPU.

\paragraph{Selection decides accuracy.}
A tuned RAGFlow pipeline answered 77 of 100. Passing the model less text helps
only when the selected text contains the answer; when it does not, no amount of
reading ability in the model can recover the fact. Long contexts are not a
substitute either, since models use them unevenly \citep{liu-lostmiddle-2024}.

\subsection{Seven real-world datasets}
\label{sec:dirty}

\paragraph{Setup.}
To test on data we had not tuned on, we drew 349 questions from seven public
or real-world sources: help-desk tickets, customer-support logs, PDFs with
extraction errors, SWE-bench Lite, The Stack, AgentBench and WebArena. The
model was Gemma~4 31B on three runtimes: llama.cpp (A6000), vLLM 0.29
(RTX 6000 Ada) and SGLang 0.5.20 (L40S). Seed 20260927; one run per runtime.

\begin{table}[t]
\centering
\small
\caption{Correct answers on seven real-world datasets (llama.cpp, Gemma~4
31B). Totals give the range across the three runtimes.}
\label{tab:dirty}
\begin{tabular}{@{}lrrr@{}}
\toprule
Dataset & Questions & Taliesin only & Taliesin + Blaise \\
\midrule
Help-desk tickets & 50 & 50 & 48 \\
Customer support & 50 & 50 & 49 \\
Messy PDFs$^{a}$ & 50 & 38 & 32 \\
SWE-bench Lite & 50 & 47 & 44 \\
The Stack & 50 & 49 & 49 \\
AgentBench & 50 & 50 & 50 \\
WebArena & 49 & 49 & 49 \\
\midrule
Total, three runtimes & 349 & 329--333 & 319--321 \\
\bottomrule
\end{tabular}

\vspace{2pt}
\begin{minipage}{0.8\linewidth}\footnotesize
$^{a}$In 9 of 50 questions the answer was lost during PDF text extraction,
before any model saw it. On the 41 answerable questions: Taliesin only 37,
Taliesin + Blaise 30.
\end{minipage}
\end{table}

\paragraph{Results.}
Across all 349 questions and three runtimes, 98.69\% of prompt tokens were
loaded from Taliesin rather than recomputed (llama.cpp 99.49\%, vLLM 99.37\%,
SGLang 97.21\%). Taliesin alone answered 329--333 questions (94--95\%).
Adding Blaise answered 319--321 (91--92\%) and was 3--10\tx{} faster per
question than Taliesin alone, on data it had never seen
(Table~\ref{tab:dirty}).

\subsection{Serving performance}

\paragraph{Time to first token.}
With Gemma~4 12B on vLLM and a prompt of about 5{,}200 tokens (7 fresh and 6
reuse requests per GPU), loading saved state was faster than recomputing on all
seven GPU types tested. The size of the gain differed by GPU: 1{,}138.8\,ms to 391.2\,ms (2.91\tx{}) on an RTX PRO
4500, and 251.2\,ms to 191.0\,ms (1.32\tx{}) on an H100 SXM.

\paragraph{Throughput under load.}
On an H100 with Qwen3-30B-A3B (16 sessions of 14{,}000 tokens, concurrency 8),
throughput rose from 2.643 to 3.424 turns per second (+29.6\%; three runs:
3.427, 3.416, 3.424). With one in three loads deliberately failed, throughput
was 2.716 turns per second, still above the baseline. Storage medium mattered little in this setup: RAM disk and local
disk gave 2.027 and 2.077 turns per second (one run each).

\paragraph{Restore latency.}
Restoring one 24{,}018-token block from local disk took 81\,ms ($\pm$0.5\,ms,
10 repeats; A40, llama.cpp); the full request including 8 generated tokens took
314\,ms.

\paragraph{Storage footprint.}
The 96{,}726-token recall corpus occupied 17.1--17.2\,GB on disk for Gemma~4
31B with llama.cpp and vLLM, and 39.0\,GB with SGLang. For DeepSeek-V4-Flash (284B),
93{,}157 tokens occupied 27.6\,GB. The store's size is capped by a configurable
disk budget with a minimum free-space floor. In the H100 throughput test
above (16 sessions, concurrency 8), RAM disk and local NVMe gave the same
throughput, so storage bandwidth was not the limit at that load.

\paragraph{Long windows at flat GPU memory.}
Because saved blocks are streamed in and discarded after use, the corpus size
is limited by storage rather than GPU memory. On an A40 with Gemma~4 and
llama.cpp, a corpus of 200 blocks of 30{,}000 tokens (5.97 million tokens)
answered 5 of 5 probes, with access time between 0.551 and 0.593\,s from depth
0 to 5.97M and GPU memory between 24{,}433 and 24{,}696\,MB. A ladder from 1M
to 50M tokens answered 24 of 25 probes (the one miss was a parsing error in the
test harness) at peak GPU memory of 33{,}813--34{,}080\,MB.

\paragraph{Kubernetes.}
On a GKE cluster with an L4 GPU, 63.9\% of lookups hit saved state
across a pod restart, and a forced kill (SIGKILL) left 0 damaged records.

\subsection{Coverage}

\paragraph{Models.}
On 4\tx{}A10G with vLLM 0.30.0 and the release library (ABI 1.31), 30 of 30
models answered, saved, loaded and hit saved state, as reported by Galahad's
own telemetry. A tensor-parallel run (Qwen3 14B, TP=2) succeeded on 5 of
5 fresh starts.

\paragraph{A 284B model.}
On DeepSeek-V4-Flash (284B) with vLLM 0.29 on 2\tx{}H200, Taliesin and Blaise
together answered 98 of 100 recall questions at 0.133\,s and 143\,J per
question.

\paragraph{Clean install.}
In a customer-style installation (Qwen3 8B), vLLM
completed one save, one load and one hit; on SGLang 0.5.20, 1{,}792 of 1{,}824
prompt tokens came from Galahad after SGLang's own cache was emptied.

\section{Design boundaries}
\label{sec:limits}

Three properties of the results follow directly from the design.

\begin{itemize}
  \item \textbf{Block-level reuse is the correct granularity.} Rotary position
  encodings make each KV row depend on its position, so individual rows cannot
  be shared between prompts (0 of 344{,}064 rows matched in a direct test).
  Galahad reuses whole blocks, where positions match exactly and the restored
  state is bit-identical.
  \item \textbf{Fail-closed by design.} Any load that fails a check is
  recomputed and the request completes. With one in three loads deliberately
  failed on an H100, throughput stayed above the baseline without Galahad
  (2.716 against 2.643 turns per second).
  \item \textbf{Memory starts from the text it is given.} In 9 of 50 messy-PDF
  questions the answer was lost by the PDF text extractor before inference. No
  memory layer can restore text that an upstream parser dropped.
\end{itemize}

\section{Related work}
\label{sec:related}

\paragraph{KV reuse inside the engine.}
PagedAttention in vLLM \citep{kwon-pagedattention-2023} and RadixAttention in
SGLang \citep{zheng-sglang-2024} share KV state between requests with a common
prefix while it remains in GPU memory. Prompt Cache
\citep{gim-promptcache-2024} reuses attention state for predefined prompt
modules. Galahad uses these engines' own interfaces and extends reuse beyond
GPU memory and process lifetime.

\paragraph{KV offloading and storage.}
LMCache \citep{liu-lmcache-2025} and Mooncake \citep{qin-mooncake-2025} move
KV state to CPU memory, disk or remote storage. CacheGen
\citep{liu-cachegen-2024} compresses KV state for transfer, and CacheBlend
\citep{yao-cacheblend-2025} and RAGCache \citep{jin-ragcache-2024} reuse KV
state for retrieved documents, accepting some approximation when chunks are
combined. Galahad differs in requiring exact reuse, with bit-identical output
as the acceptance test, and in treating any mismatch as a miss.

\paragraph{Retrieval.}
Retrieval-augmented generation \citep{lewis-rag-2020} passes selected text to
the model instead of the whole corpus; lexical ranking such as BM25
\citep{robertson-bm25-2009} remains a strong baseline. Blaise is a retrieval
component in this sense. Its design goal is to return exact document sections rather than similarity-ranked chunks.

\paragraph{Compact context.}
Cartridges \citep{eyuboglu-cartridges-2025} train a compact KV representation
of a corpus. Galahad stores the model's unmodified KV state and requires no
training.

\paragraph{Prior work by the author.}
Galahad builds on Merlin, a byte-exact deduplication engine
\citep{schelpe-merlin-lossless-2026,schelpe-rag-dedup-2026}, and on earlier
results on byte-exact KV grafting and long windows at flat GPU memory
\citep{schelpe-grafting-2026,schelpe-frozen12b-2026}. This paper reports the
integrated system and its first evaluation across three runtimes.

\section{Discussion}
\label{sec:discussion}

The results support three claims. First, a model's
reading can be stored and reused exactly: restored state produced
bit-identical logits, and the approach worked on 30 of 30 tested models.
Second, reuse turns most prompt computation into loading: on seven real-world
datasets, 98.7\% of prompt tokens came from memory. Third, each memory helps on its own: KV
persistence alone let a model answer from a corpus eight times larger than its
window, 3.1\tx{} faster than the truncated baseline on llama.cpp, and adding the
text memory cut time by a further 4.7\tx{} and energy by a further 2.7\tx{}.

These results change what inference costs. Today
the cost of a request grows with the total text in its prompt. With a
persistent memory, it grows with the text the model has not read before. For
workloads that return to the same documents, such as support, code, legal and
agent tasks, this cost is much smaller. It also changes which models are
practical: a smaller model with exact memory of a large corpus can answer
questions that would otherwise require a model with a much longer context
window. This shift from stateless to stateful inference is the main
implication of the work.

\paragraph{Relation to the compute ceiling.}
Galahad does not contradict the bound of Sikka and Sikka
\citep{sikka-hallucination-2025}. The model that answers is unchanged, and so
is the computation it performs per token. What changes is what that
computation is spent on, in three ways. First, work is done once: attention
state computed for a block is stored exactly and loaded on later requests, so
the model's budget is not spent recomputing text it has already read. Second,
search moves outside the model: finding the relevant section in a corpus, the
part of the task whose cost grows with the corpus, runs on the CPU, and the
model receives only the task of reading one section and answering. Third,
verification moves outside the model: whether loaded state is correct is
decided by byte comparison and cryptographic hashes, not by the model judging
its own output, which is the kind of self-verification that their argument
says fails above the bound. The ceiling stays where it is; Galahad works around it by keeping
repeated work, search and verification away from the model.

Next steps are benchmarking Blaise's second mode, in which the model reads the
corpus index once and Taliesin keeps that reading; extending section
selection to poorly extracted PDFs; and results from more models and
independent operators.

\section{Availability}
\label{sec:availability}

Galahad enters public beta on 1 October 2026 at
\url{https://github.com/corbenicai/galahad}. The beta is free for
non-commercial use on one GPU for 12 months, renewable; commercial pilots are
available on request. Raw logs for the tables in this paper will be published
with the accompanying dataset. Patent applications covering parts of the
system are pending.

\section*{Disclosure}
The author is the founder of Corbenic AI, which develops Galahad. All
experiments were run by the author on rented cloud GPUs.

\bibliographystyle{plainnat}
\bibliography{references}

@article{schelpe-merlin-lossless-2026,
  author  = {Sietse Schelpe},
  title   = {{Merlin}: Deterministic Byte-Exact Deduplication for Lossless Context Optimization in Large Language Model Inference},
  journal = {arXiv preprint arXiv:2605.09990},
  year    = {2026}
}

@article{schelpe-rag-dedup-2026,
  author  = {Sietse Schelpe},
  title   = {Byte-Exact Deduplication in Retrieval-Augmented Generation: A Three-Regime Empirical Analysis Across Public Benchmarks},
  journal = {arXiv preprint arXiv:2605.09611},
  year    = {2026}
}

@article{schelpe-grafting-2026,
  author  = {Sietse Schelpe},
  title   = {Smarter and Cheaper at Once: Byte-Exact {KV}-Cache Grafting Turns a Frozen Small Model into a Verified-Knowledge Flywheel},
  journal = {arXiv preprint arXiv:2607.14431},
  year    = {2026}
}

@article{schelpe-frozen12b-2026,
  author  = {Sietse Schelpe},
  title   = {A Frozen {12B} Beats Frontier Models on Verified Work: 100\% Accuracy, 0 Tokens, Bit-Exact, Forever},
  journal = {arXiv preprint arXiv:2607.23806},
  year    = {2026}
}

@inproceedings{kwon-pagedattention-2023,
  author    = {Woosuk Kwon and Zhuohan Li and Siyuan Zhuang and Ying Sheng and Lianmin Zheng and Cody Hao Yu and Joseph E. Gonzalez and Hao Zhang and Ion Stoica},
  title     = {Efficient Memory Management for Large Language Model Serving with {PagedAttention}},
  booktitle = {Proc. SOSP},
  year      = {2023},
  note      = {arXiv:2309.06180}
}

@inproceedings{zheng-sglang-2024,
  author    = {Lianmin Zheng and Liangsheng Yin and Zhiqiang Xie and Chuyue Sun and Jeff Huang and Cody Hao Yu and Shiyi Cao and Christos Kozyrakis and Ion Stoica and Joseph E. Gonzalez and Clark Barrett and Ying Sheng},
  title     = {{SGLang}: Efficient Execution of Structured Language Model Programs},
  booktitle = {Proc. NeurIPS},
  year      = {2024},
  note      = {arXiv:2312.07104}
}

@article{liu-lmcache-2025,
  author  = {Yuhan Liu and Yihua Cheng and Jiayi Yao and Yuwei An and Xiaokun Chen and Shaoting Feng and Yuyang Huang and Samuel Shen and Rui Zhang and Kuntai Du and Junchen Jiang},
  title   = {{LMCache}: An Efficient {KV} Cache Layer for Enterprise-Scale {LLM} Inference},
  journal = {arXiv preprint arXiv:2510.09665},
  year    = {2025}
}

@inproceedings{qin-mooncake-2025,
  author    = {Ruoyu Qin and Zheming Li and Weiran He and Mingxing Zhang and Yongwei Wu and Weimin Zheng and Xinran Xu},
  title     = {{Mooncake}: Trading More Storage for Less Computation -- A {KVCache}-centric Architecture for Serving {LLM} Chatbot},
  booktitle = {Proc. FAST},
  year      = {2025},
  note      = {arXiv:2407.00079}
}

@inproceedings{liu-cachegen-2024,
  author    = {Yuhan Liu and Hanchen Li and Yihua Cheng and Siddhant Ray and Yuyang Huang and Qizheng Zhang and Kuntai Du and Jiayi Yao and Shan Lu and Ganesh Ananthanarayanan and Michael Maire and Henry Hoffmann and Ari Holtzman and Junchen Jiang},
  title     = {{CacheGen}: {KV} Cache Compression and Streaming for Fast Large Language Model Serving},
  booktitle = {Proc. SIGCOMM},
  year      = {2024},
  note      = {arXiv:2310.07240}
}

@inproceedings{yao-cacheblend-2025,
  author    = {Jiayi Yao and Hanchen Li and Yuhan Liu and Siddhant Ray and Yihua Cheng and Qizheng Zhang and Kuntai Du and Shan Lu and Junchen Jiang},
  title     = {{CacheBlend}: Fast Large Language Model Serving for {RAG} with Cached Knowledge Fusion},
  booktitle = {Proc. EuroSys},
  year      = {2025},
  note      = {arXiv:2405.16444}
}

@inproceedings{gim-promptcache-2024,
  author    = {In Gim and Guojun Chen and Seung-seob Lee and Nikhil Sarda and Anurag Khandelwal and Lin Zhong},
  title     = {Prompt {Cache}: Modular Attention Reuse for Low-Latency Inference},
  booktitle = {Proc. MLSys},
  year      = {2024},
  note      = {arXiv:2311.04934}
}

@article{jin-ragcache-2024,
  author  = {Chao Jin and Zili Zhang and Xuanlin Jiang and Fangyue Liu and Xin Liu and Xuanzhe Liu and Xin Jin},
  title   = {{RAGCache}: Efficient Knowledge Caching for Retrieval-Augmented Generation},
  journal = {arXiv preprint arXiv:2404.12457},
  year    = {2024}
}

@article{eyuboglu-cartridges-2025,
  author  = {Sabri Eyuboglu and Ryan Ehrlich and Simran Arora and Neel Guha and Dylan Zinsley and Emily Liu and Will Tennien and Atri Rudra and James Zou and Azalia Mirhoseini and Christopher R{\'e}},
  title   = {Cartridges: Lightweight and General-Purpose Long Context Representations via Self-Study},
  journal = {arXiv preprint arXiv:2506.06266},
  year    = {2025}
}

@inproceedings{lewis-rag-2020,
  author    = {Patrick Lewis and Ethan Perez and Aleksandra Piktus and Fabio Petroni and Vladimir Karpukhin and Naman Goyal and Heinrich K{\"u}ttler and Mike Lewis and Wen-tau Yih and Tim Rockt{\"a}schel and Sebastian Riedel and Douwe Kiela},
  title     = {Retrieval-Augmented Generation for Knowledge-Intensive {NLP} Tasks},
  booktitle = {Proc. NeurIPS},
  year      = {2020},
  note      = {arXiv:2005.11401}
}

@article{robertson-bm25-2009,
  author  = {Stephen Robertson and Hugo Zaragoza},
  title   = {The Probabilistic Relevance Framework: {BM25} and Beyond},
  journal = {Foundations and Trends in Information Retrieval},
  volume  = {3},
  number  = {4},
  pages   = {333--389},
  year    = {2009}
}

@article{liu-lostmiddle-2024,
  author  = {Nelson F. Liu and Kevin Lin and John Hewitt and Ashwin Paranjape and Michele Bevilacqua and Fabio Petroni and Percy Liang},
  title   = {Lost in the Middle: How Language Models Use Long Contexts},
  journal = {Transactions of the Association for Computational Linguistics},
  volume  = {12},
  pages   = {157--173},
  year    = {2024},
  note    = {arXiv:2307.03172}
}

@misc{ragflow-2026,
  author       = {{InfiniFlow}},
  title        = {{RAGFlow}: open-source retrieval-augmented generation engine, version 0.27.2},
  year         = {2026},
  howpublished = {\url{https://github.com/infiniflow/ragflow}}
}

@misc{he-nondeterminism-2025,
  author       = {Horace He and {Thinking Machines Lab}},
  title        = {Defeating Nondeterminism in {LLM} Inference},
  year         = {2025},
  howpublished = {\url{https://thinkingmachines.ai/blog/defeating-nondeterminism-in-llm-inference/}}
}

@misc{vllm-issue-33123-2025,
  author       = {{AndreasKaratzas}},
  title        = {[{Bug}][{ROCm}]: Prefix caching produces different output on first request (cache miss) vs subsequent requests (cache hit)},
  year         = {2026},
  howpublished = {vLLM GitHub issue \#33123, opened 26 January 2026, \url{https://github.com/vllm-project/vllm/issues/33123}}
}

@article{sikka-hallucination-2025,
  author  = {Varin Sikka and Vishal Sikka},
  title   = {Hallucination Stations: On Some Basic Limitations of Transformer-Based Language Models},
  journal = {arXiv preprint arXiv:2507.07505},
  year    = {2025}
}

\end{document}